\documentclass[runningheads]{llncs}
\usepackage{graphicx}
\usepackage{tikz}
\usepackage{comment}
\usepackage{amsmath,amssymb} % define this before the line numbering.
\usepackage{color}

\usepackage{booktabs}
\usepackage[pagebackref,breaklinks,colorlinks]{hyperref}

\usepackage{epsfig}
\usepackage{graphicx}
\RequirePackage{xspace}

\usepackage{nicefrac}
\usepackage{siunitx}
\usepackage[symbol]{footmisc}
\usepackage{multirow}
\usepackage{array}
\newcolumntype{H}{>{\setbox0=\hbox\bgroup}c<{\egroup}@{}}  % Allows to hide a table column with H

\newcommand{\squishlist}{
	\begin{list}{$\bullet$}
		{ \setlength{\itemsep}{0pt}
			\setlength{\parsep}{1pt}
			\setlength{\topsep}{1pt}
			\setlength{\partopsep}{0pt}
			\setlength{\leftmargin}{1.5em}
			\setlength{\labelwidth}{1em}
			\setlength{\labelsep}{0.5em} } }
\newcommand{\squishend}{\end{list} 
}

\DeclareMathOperator*{\argmin}{arg\,min}

\usepackage[capitalize]{cleveref}
\usepackage{babel}
\crefname{section}{Sec.}{Secs.}
\Crefname{section}{Section}{Sections}
\Crefname{table}{Table}{Tables}
\crefname{table}{Tab.}{Tabs.}

\makeatletter
\DeclareRobustCommand\onedot{\futurelet\@let@token\@onedot}
\def\@onedot{\ifx\@let@token.\else.\null\fi\xspace}

\def\eg{\emph{e.g}\onedot}

\makeatother

\usepackage[accsupp]{axessibility}  % Improves PDF readability for those with disabilities.

\begin{document}
% \linenumbers
\pagestyle{headings}
\mainmatter
\def\ECCVSubNumber{6177}  % Insert your submission number here

\title{3D Scene Inference from Transient Histograms}

\begin{comment}
\titlerunning{ECCV-22 submission ID \ECCVSubNumber} 
\authorrunning{ECCV-22 submission ID \ECCVSubNumber} 
\author{Anonymous ECCV submission}
\institute{Paper ID \ECCVSubNumber}
\end{comment}
%******************

% \begin{comment}
\titlerunning{HistoVision}
% \author{First Author\inst{1}\orcidID{0000-1111-2222-3333} \and Second Author\inst{2,3}\orcidID{1111-2222-3333-4444} \and Third Author\inst{3}\orcidID{2222--3333-4444-5555}}
\author{Sacha Jungerman \and Atul Ingle \and Yin Li \and Mohit Gupta}
\authorrunning{S. Jungerman et al.}
\institute{University of Wisconsin-Madison, Madison WI 53706, USA\\
\email{\{sjungerman, ingle, yin.li, mgupta37\}@wisc.edu}}
% \end{comment}
%******************
\maketitle

%%%%%%%%% ABSTRACT

%With advancements in single-photon sensitive image sensors like single-photon avalanche diodes (SPADs) it is now possible to record light at these short time scales at low cost and low power.

\begin{abstract}
  Time-resolved image sensors that capture light at pico-to-nanosecond timescales were once limited to niche applications but are now rapidly becoming mainstream in consumer devices. We propose low-cost and low-power imaging modalities that capture scene information from minimal time-resolved image sensors with as few as one pixel. The key idea is to flood illuminate large scene patches (or the entire scene) with a pulsed light source and measure the time-resolved reflected light by integrating over the entire illuminated area. The one-dimensional measured temporal waveform, called \emph{transient}, encodes both distances and albedoes at all visible scene points and as such is an aggregate proxy for the scene's 3D geometry. We explore the viability and limitations of the transient waveforms by themselves for recovering scene information, and also when combined with traditional RGB cameras. We show that plane estimation can be performed from a single transient and that using only a few more it is possible to recover a depth map of the whole scene. We also show two proof-of-concept hardware prototypes that demonstrate the feasibility of our approach for compact, mobile, and budget-limited applications.
  
  %\vspace{-1em}

%that is significantly more accurate than that of existing methods

%and accompanying integrated pulsed lasers

%  To exploit these, we design a deep network architecture, dubbed
%  YNet, that takes a transient histogram along with a regular RGB image to
%  recover a dense depth map.

%   Our simulation results show that our algorithm
%   outperforms the state-of-the-art learning-based monocular depth estimation by
%   a significant margin.  We also show two proof-of-concept hardware prototypes that
%   show the feasibility of our approach for compact, mobile, and budget-limited applications.

%  We set out to perform monocular depth estimation with high accuracy given
%  minimal extra information about the scene. We chose to use a transient
%  histogram, provided by a time-resolved single-photon sensor such as a
%  single-photon avalanche diode (SPAD), as this extra information. Using this,
%  we developed a network architecture, dubbed YNet, that takes this extra
%  information into account and outperforms traditional UNet based depth
%  estimation techniques.
\end{abstract}

%%%%%%%%% BODY TEXT
%\vspace{-1em}
\section{Weak 3D Cameras}\label{sec:departure}

\begin{figure*}[!ht]
    \centering
    \includegraphics[width=1.0\textwidth]{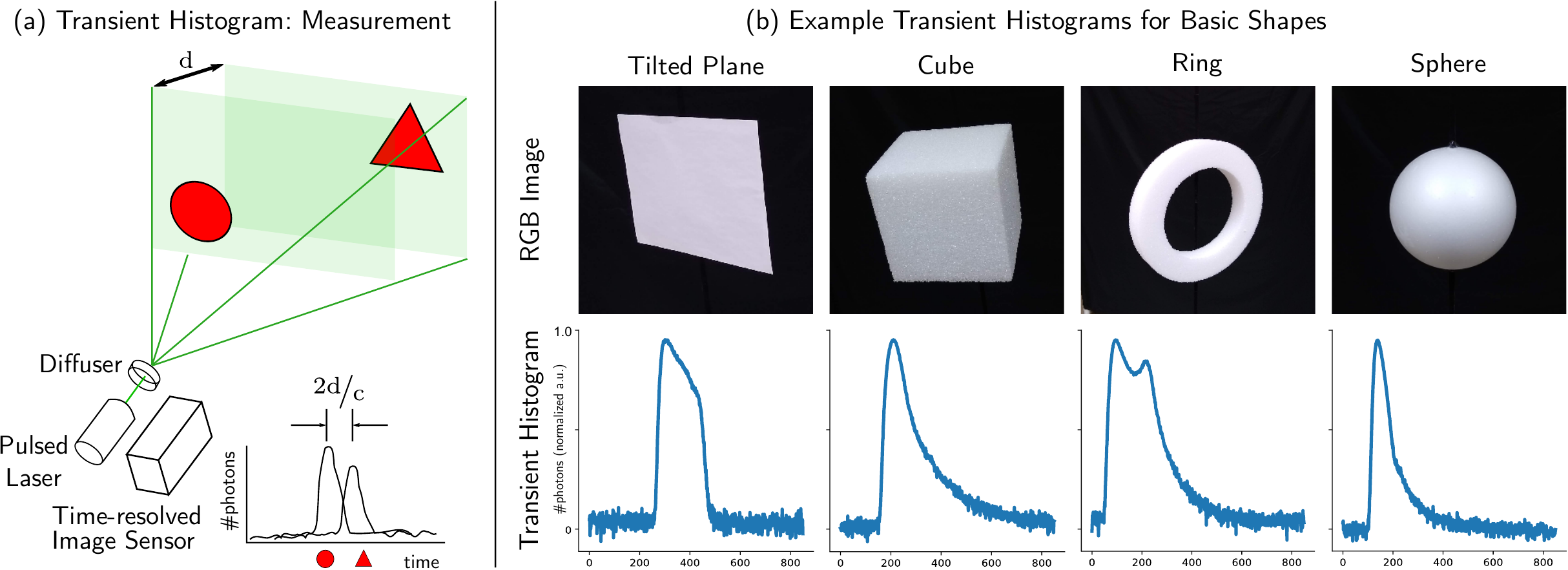}
    \caption{\textbf{Transient histogram: Measurement and examples.} (a) Measuring a
    transient histogram involves illuminating the scene with a pulsed illumination
    source such as a pulsed laser which is diffused uniformly over the field of view
    of a time-resolved image sensor pixel such as a single-photon avalanche diode.
    (b) Transient histograms were captured using a hardware prototype for some
    basic 3D shapes. Observe that these have unique features which can help distinguish these
    shapes.\label{fig:teaser}}
    %\vspace{-1.5em}
\end{figure*}

Vision and robotics systems enabled by 3D cameras are revolutionizing several aspects of our lives via technologies such as robotic surgery, augmented reality, and autonomous navigation. One catalyst behind this revolution is the emergence of depth sensors that can recover the 3D geometry of their surroundings. While a full 3D map may be needed for several applications such as industrial inspection and digital modeling, there are many scenarios where recovering high-resolution 3D geometry is not required. Imagine a robot delivering food on a college campus or a robot arm sorting packages in a warehouse. In these settings, while full 3D perception may be useful for long-term policy design, it is often unnecessary for time-critical tasks such as obstacle avoidance. There is strong evidence that many biological navigation systems such as human drivers~\cite{lee_theory_1976} do not explicitly recover full 3D geometry for making fast, local decisions such as collision avoidance. For such applications, particularly in resource-constrained scenarios where the vision system is operating under a tight budget (e.g., low-power, low-cost), it is desirable to have \emph{weak 3D cameras} which recover possibly low-fidelity 3D scene representations, but with low latency and limited power.

We propose a class of weak 3D cameras based on \emph{transient histograms}, a scene representation tailored for time-critical and resource-constrained applications such as fast robot navigation. A transient histogram is a one-dimensional signal (as opposed to 2D images) that can be captured at high speeds and low-costs by re-purposing cheap proximity sensors that are now ubiquitous, everywhere from consumer electronics such as mobile phones, to cars, factories, and robots for collision safety. Most proximity sensors consist of a laser source and a fast detector and are based on the principle of time-of-flight (ToF): measuring the time it takes for a light pulse to travel to the scene and back to the sensor.

Conventionally, in a ToF sensor, both the fields of view (FoV) of the laser and that of the detector need to coincide and be highly focused (ideally at a single scene point). This ensures that the received light intensity has a single discernible peak corresponding to the round-trip time delay. We adopt a different approach. Instead of focusing the beam on a narrow region, we deliberately \emph{diffuse} both the laser and the detector so that a large scene area is illuminated simultaneously. The received light is composed of the superposition of all scaled and shifted light pulses from all the illuminated scene points. The resulting captured waveform is called the transient histogram or simply a transient. Instead of encoding the depth of a single scene point, a transient is an \emph{aggregate 3D scene representation} that encodes information about the 3D geometry and albedos of a large scene patch, possibly even the entire scene.

%It is possible to capture the transients with low-cost proximity sensors simply by diffusing their FoVs and tweaking their software.

%For these scenes, we show that plane parameters can be analytically recovered as well.

We propose a family of algorithms that can extract scene information from transients, beyond what is achieved simply via peak-finding. These methods broadly fall under two categories: parametric and non-parametric. For scenes where some prior knowledge can be explicitly modeled, we show an analysis-by-synthesis approach that can be used to recover the scene parameters. While this technique can be used for arbitrary parametric scenes, we show results for planar scenes in Section~\ref{sec:plane-est}. We demonstrate the viability of plane estimation using a hardware prototype that uses a low-cost, off-the-shelf proximity sensor. Finally, for more complicated scenes, we present a learning-based method in Section~\ref{sec:depth-est} to recover a dense depth map using only a small (e.g., $20 \times 15$) array of transients. We then show how these depth maps can be further refined using an RGB image, and demonstrate these techniques on a custom hardware setup which is more flexible than a cheap off-the-shelf sensor, yet mimics its characteristics closely.

\smallskip
\noindent {\bf Scope and Limitations:} The proposed techniques are specifically tailored for applications running on resource-constrained devices, where a low-fidelity depth representation suffices. Our methods are not meant to replace conventional depth cameras in scenarios that require dense and high-precision geometry information. Instead, the transient histograms should be considered a complementary scene representation that can be recovered with low latency and compute budgets using only low-cost proximity sensors.

% Yin: this reads like a summary of the paper rather than a description of limitations. 
Due to an inherent rotational ambiguity when estimating planes from a single transient, an analytic solution to recover depth of a piecewise planar scene is infeasible. Instead, we adopt a deep learning approach to estimate the geometry of complex scenes. Despite successful demonstration of plane estimation results with a cheap proximity sensor, using SPAD sensors to image more complex scenes is still challenging due to data bandwidth, low signal-to-noise ratio (SNR), and range issues. We instead show depth estimation results using a custom lab setup that was built to perform similarly as off-the-shelf proximity sensors, while providing us with greater flexibility and low-level access to transient histograms.

\section{Related Work}\label{sec:related-work}

%Traditionally to estimate scene geometry, scene depth needed to be acquired. Consequently, d

% Why not use a more classical approach?
\noindent {\bf 3D Imaging Techniques}: Traditional depth sensing methods such as stereo, structured-light, and time-of-flight are capable of acquiring high-quality depth maps. While these methods have made significant progress, such as multi-camera depth estimation~\cite{zhang_domain-invariant_2019,wang_pseudo-lidar_2020,zhang_depth_2020}, they still face key challenges. Certain applications such as autonomous drones, cannot support complex structured-light sensors, multiple cameras, or bulky LiDARs due to cost, power, and other constraints. We propose a method suitable for budget-constrained applications
which is less resource-intensive because it can estimate depth maps from just a few transients.

% What about MDE?

%Estimating depth from a single image is fundamentally ambiguous, as the same 2D projection might have several equally valid 3D explanations.

%to regress a dense depth map from an input RGB image. For example,
%can be used to recover the fine details of the depth maps

\smallskip
\noindent {\bf Monocular Depth Estimation (MDE)}: A promising low-cost depth recovery technique is monocular depth estimation which aims to estimate dense depth maps from a single RGB image. Early works on MDE focus on hand-crafted appearance features~\cite{hoiem_automatic_2005,hoiem_recovering_2007,saxena_learning_2005,saxena_make3d_2009} such as color, position, and texture. Modern MDE techniques use almost exclusively learning-based approaches, including multi-scale deep networks~\cite{eigen_predicting_2015,liu_learning_2016}, attention mechanisms~\cite{hao_detail_2018,aich_bidirectional_2020,huynh_guiding_2020}, and recently vision-transformers~\cite{ranftl_vision_2021}. Despite predicting ordinal depth well and providing exceptional detail, existing methods cannot resolve the inherent scale ambiguity, resulting in overall low depth accuracy as compared to LiDAR systems.

%a few modern approaches have expanded upon the problem and added external information to help guide the prediction. A
%Others have taken a probabilistic approach to MDE~\cite{xia_generating_2019} which does not predict a depth map, but a distribution of depths. This distribution then allows for sampling or conditioning on any extra information that we might have, like sparse depth, average depth, etc.
%Moreover, using external information to guide MDE needs to acquire said extra data through additional sensing apparatus. A cost-effective way to obtain this extra information is to use a proximity sensor that can produce a transient histogram.

% Augment MDE?
One possible approach to overcome this ambiguity is to augment the input RGB image with additional information such as the scene's average depth~\cite{zhou_unsupervised_2017,alhashim_high_2019} or a sparse depth map~\cite{xia_generating_2019,bergman_deep_2020}. Some approaches have also designed specialized optics that help disambiguate depth~\cite{chang_deep_2019,wu_phasecam3d_2019}. However, these approaches either require customized hardware that is not readily available or rely on some external information that cannot easily be procured. Sensor fusion techniques like that of Lindell et al.~\cite{lindell_single-photon_2018} learn to predict depth maps from dense $256 \times 256$ ToF measurements and a collocated RGB image. In contrast, our proposed method does not need an RGB image, and yields good results with as few as $4\times3$ transients, thus saving on power and compute requirements. 

%While this method produces good results, it does so only if the input depth estimate is ordinally correct.

% Brief prior art with transients. What if no RGB?
% Others have forgone the RGB camera completely. Instead of using spatial information, such as an RGB image, to predict scene geometry we can only use temporal data such as that contained in a transient.
\smallskip
\noindent {\bf Transient-based Scene Recovery}: Recently, transient-based depth recovery methods have been proposed, including a two-step process~\cite{nishimura_disambiguating_2020} that first uses a pretrained MDE network to predict a rough depth estimate which then gets tuned as its transient gets aligned to match a captured one. This approach relies on the original MDE-based depth map to be ordinally correct for achieving high-quality depths. Further, this two-step method can only work in the presence of a collocated RGB camera. Callenberg et al.~\cite{callenberg_low-cost_2021} use a 2-axis galvo-mirror system with a low-cost proximity sensor to scan the scene. This leads to an impractical acquisition time of more than 30 minutes even for a relatively low $128\times128$ resolution scan. Our goal is different. We aim to recover 3D scene information from a single transient or a sparse spatial grid (e.g., $20 \times 15$) of transients, with minimal acquisition times and computation costs.

\smallskip
\noindent\textbf{Non-Line-of-Sight Imaging (NLOS)}: 
NLOS techniques aim to recover hidden geometry using indirect reflections from occluded objects~\cite{pediredla_reconstructing_2017,metzler_keyhole_2021,tsai_geometry_2017,xin_theory_2019}. Instead of diffusing light
off a relay surface, our method diffuses the source light in a controlled manner and only captures direct scene reflections. This provides higher SNR, enables use of off-the-shelf detectors, and consumes lower power by estimating depth with orders of magnitude fewer transients than NLOS methods.

\section{Transient Histograms}
% \vspace{-2em}
%TODO show radiometric image formation here
% no SPAD or histogram needed for this, follow a ray to and from a scene patch!
% reuse equations from here: https://jungerm2.github.io/HistoVision/#physical-sensor-modeling
% by the end of section 3 we have a model for r_i
%
% then talk about what information is contained in these transient histograms. in a new section

We consider an active imaging system that flash illuminates
a large scene area by a periodic pulse train of Dirac delta laser
pulses, where each pulse deposits $\Phi_\text{laser}$
photons into the scene over a fixed exposure time. The laser light is uniformly
diffused over an illumination cone angle $\theta$. In
practice, this can be achieved with a diffuser as in Fig.\
\ref{fig:teaser}(a).  Let $f$ denote the repetition
frequency of the laser. The unambiguous depth range
is given by $r_\text{max} = \frac{c}{2f}$ where $c$ is the
speed of light.

The imaging system also includes a single-pixel lens-less time-resolved image sensor, co-located with the laser. The sensor collects photons returning from the field of view illuminated by the laser source. Let $\Delta$ be the time resolution of the image sensor pixel, which corresponds to a distance resolution of $\frac{c\Delta}{2}$. The unambiguous depth range $r_\text{max}$ is therefore discretized into $N = \lfloor \nicefrac{1}{f\Delta} \rfloor$ bins.

\subsection{Image Formation}\label{sec:tdh-formation}
\begin{figure}[!t]
    \centering
    \includegraphics[width=0.95\columnwidth]{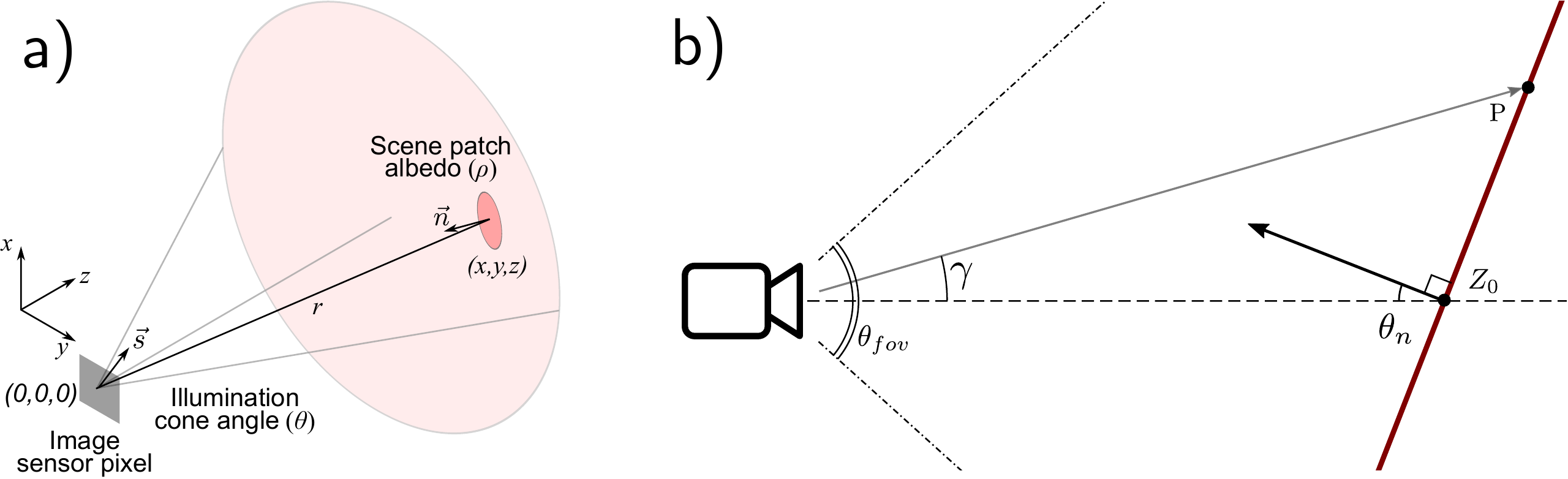}
    \caption{\textbf{Geometry for radiometric image formation model and planar parametrization.} (a) The single-pixel image sensor receives light from a scene patch with albedo $\rho$ located at a distance $r$.  The normal vectors at the scene patch and the sensor pixel are denoted by $\vec{n}$ and $\vec{s}$. The transient response at a fixed time delay is computed by integrating the total light returning from all scene patches located in a small range corresponding to the time resolution of the sensor pixel. (b) When viewed in the $(\vec{n}, \vec{s})$ plane, the scene plane is parametrized by its distance $Z_0$ from the sensor and its angle $\theta_n$ measured with respect to sensor's optical axis.
    \label{fig:scene-setup-radiometry}}
    %\vspace{-1.5em}
\end{figure}

Fig.\ \ref{fig:scene-setup-radiometry}(a) shows the imaging
geometry used for deriving the radiometric image formation
model, where a 3D coordinate system is fixed with the
single-pixel sensor at the origin and positive $z$-axis
pointing into the scene. The laser source is also located at
the origin with its direction defined by $\vec{s}$. We
assume that the scene is composed of a discrete collection 
of perfectly Lambertian scene patches.
Each visible scene patch has a depth $z$
parametrized by its $(x,y)$
location. Thus, the albedo and surface normal of each patch
is given by $\rho(x,y)$ and $\vec{n}(x, y)$. We assume that
there are no inter-reflections within the scene and all
scene patches that contribute to the signal received by the
sensor are inside the unambiguous depth range $0\!\leq\!
r\!:=\!\sqrt{x^2+y^2+[z(x,y)]^2}\! <\! r_\text{max} =
\nicefrac{c}{2f}.$

The received laser photon flux vector consists of $N$ bins with mean rates given by $\boldsymbol{\varphi} = (\varphi_1, \varphi_2, \ldots, \varphi_N)$. We call $\boldsymbol{\varphi}$ the \textbf{transient histogram}. The photon flux $\varphi_i^\text{laser}$ contributed by the laser illumination at the $i^\text{th}$ bin is given by integrating the light returning from all scene patches that lie in a range of distances satisfying $\nicefrac{(i-1)c\Delta}{2} \leq r <\nicefrac{i c\Delta}{2}$. Ignoring multi-bounce paths,
\begin{equation}
  \begin{split}
  \varphi^\text{laser}_i &= \iint_{(x,y): \frac{(i-1)c\Delta}{2} \leq r < \frac{ic\Delta}{2}}
              \frac{\rho(x,y)}{4\pi^2(1-\cos(\theta_\text{fov}/2)) r^4} (\vec{n}(x,y) \cdot \vec{s})
              \Phi_\text{laser} \,\, dr \nonumber \\
              &= \iint_{(x,y): \frac{(i-1)c\Delta}{2} \leq r < \frac{ic\Delta}{2}}
              \frac{\hat\rho(x,y)}{4\pi^2(1-\cos(\theta_\text{fov}/2)) r^4} \Phi_\text{laser} \,\, dr
  \end{split}\label{eq:signalflux}
\end{equation}
where $\hat \rho$ is the cosine-angle adjusted albedo of the scene patch. Again, $\theta_\text{fov}$ is the illumination cone angle, $\vec{n}(x,y)$ is the surface normal of the scene patch, $\vec{s}$ is the source normal and $\Phi_\text{laser}$ is the number of photons in each laser pulse.

The final transient histogram at bin $i$ is thus given by:
\begin{equation}
  \varphi_i = \varphi_i^\text{laser} + \varphi^\text{bkg}
              \label{eq:transient_histogram}
\end{equation}
where the constant background component $\varphi^\text{bkg}$ consists of the ambient photon flux (\eg, sunlight) and internal sensor noise (\eg, due to dark current). A transient histogram\footnote{This is different from a \emph{transient scene response}~\cite{otoole_reconstructing_2017} which is acquired at each $(x,y)$ location (either by raster scanning, or a sensor pixel array) whereas a transient histogram integrates over all patches.} thus forms a scene representation that integrates the effects of scene depth, surface normal, and surface albedo into a one dimensional signal, further affected by ambient light and sensor noise. When measuring a transient histogram, a random number of photons will be incident during each bin $i$ according to a Poisson distribution with mean $\varphi_i$.

\subsection{Measuring a Transient Histogram}\label{sec:tdh-measurement}
% we can measure APD or let's use SPADs! why SPADs? time res, manufacturability
% here we talk about SPAD specific image formation model how r_i translates to
% a histogram measured by a SPAD
% describe the conventional image formation, then just say we actually use asynchronous
% async acq - just cite iccv paper.
% later talk about pileup issues, maybe show effect of pileup with simulation
% show a photo of setup that captures it.
The key strength of a transient histogram is its high
temporal resolution. Such a scene representation can be captured using fast detectors that
can operate on nano-to-picosecond timescales. 
One such technology is the avalanche photodiode (APD). APDs equipped with a high-sampling-rate analog-to-digital converter can be used to capture the full transient histogram in a single laser pulse. Our hardware prototypes use a different image sensing technology called single-photon avalanche diode
(SPAD). SPADs have gained popularity in recent years due to
their single-photon sensitivity and extreme time resolution ($\sim$\SI{100}{\pico\second}).
Unlike APDs, SPAD arrays can be manufactured cheaply and at scale using standard CMOS
fabrication technology. 

\smallskip
\noindent \textbf{Estimating Transient Histograms using
SPADs:} Unlike a conventional image sensor, a SPAD pixel can
capture at most one returning photon per laser period. This
is because, after each photon detection event, the SPAD
pixel enters a \emph{dead-time} during which the pixel is
reset. Conventionally, a SPAD pixel is operated in
synchronization with the pulsed laser; photon timestamps are
acquired over many laser cycles, and a histogram of photon
counts is constructed. We call this the \emph{SPAD
histogram}. We now show that the scene's transient histogram
can be estimated from a measured SPAD histogram.

In each laser cycle, the probability $q_i$ that at least one
photon is incident on the SPAD pixel in bin $i$ can be
calculated using Poisson statistics\footnote{The quantum
efficiency of the SPAD pixel is absorbed into $\varphi_i$.}:
$q_i\!=\!  1\!-\!e^{-\varphi_i}$. The probability $p_i$ that
the SPAD captures a photon in the $i^\text{th}$ bin follows
a geometric distribution: $p_i = q_i \prod_{j<i} q_j$.  For
convenience, the $(N\!+\!1)^\text{th}$ SPAD histogram bin
stores the number of laser cycles with no photon detection:
$p_{N+1}=1-\sum_{i=1}^N p_i$. If the total incident photon
flux is low such that only a small fraction of the laser
cycles lead to a photon detection\footnote{In case of high ambient
illumination, existing pile-up mitigation techniques
\cite{gupta_asynchronous_2019,pediredla_signal_2018,rapp_high-flux_2021} can be employed.},
the expected number of photons measured by the SPAD in bin
$1\leq i \leq N$ is proportional to the transient histogram:
$\mathbf{E}[h_i] = L\varphi_i$ where $L$ is the number of
laser cycles. The transient histogram is thus $\widetilde
\varphi_i = \nicefrac{h_i}{L}$. We assume that the
SPAD pixel captures 512 bins over a \SI{10}{\m} range
corresponding to a time bin resolution of
\SI{130}{\pico\second}.

\subsection{Information in a Transient Histogram}\label{sec:tdh-info}
% there is a mix of depth and albedo information
% point to Fig 1 teaser shapes have different tdh shapes
% but there are ambiguities
% how do we recover depths from a single histogram?
% there have been attempts before but it is severely underconstrained
% and ill-posed inverse problem so we need some prior knowledge.
% use MDH!
% unlike MDH, we think of histograms as the primary source of information
% and the RGB provides side information.
% show Fig. 2 here: DNN architecture high level

A key question arises: ``What information does a transient histogram contain?''. Fig.\ \ref{fig:teaser}(b) shows example histograms for simple shapes, captured experimentally\footnote{Details of experimental hardware are discussed later in Section~\ref{subsec:lab-setup}.}. These histograms have unique features for different shapes. Each histogram has a sharp leading edge corresponding to the shortest distance from the sensor to the object. For a 2D tilted plane, the transient histogram also has a sharp trailing edge with a drop-off to zero. The support (width of the non-zero region) reveals the difference between the farthest and the nearest visible point on the object. For 3D shapes like the cube and the sphere, there is no sharp trailing edge, and the drop-off is more gradual. The 3D ring has a double peak, the distance between these peaks is a function of the angle of the plane of the ring with respect to the sensor.

%It is tempting to pose the depth recovery problem as a regularized optimization. Given a measured transient histogram $(\widetilde \varphi_i)_{i=1}^N$,
%\[
%\underset{z(x,y)}{\mbox{minimize}}\sum_{i=1}^{N} ||\varphi_i - \widetilde\varphi_i|| + R(z)
%\]
%where the first term forces the measurement to match the theoretical transient histogram $\varphi_i$ in  Eq.\ \ref{eq:transient_histogram}, and the second term, $R(z),$ is a regularization term on the depth map. For instance, to force the depth map to be piecewise constant, $R(z)$ can be a total-variation norm. %TODO cite total variation?

While the leading edge of a transient histogram gives an accurate estimate of the distance to the nearest point on an object, recovering the depth map $z(x,y)$ from a histogram is severely under-determined even for very simple shapes, as a transient histogram is an integration of depth, surface normal and albedo. Physically plausible scenes with different depth maps can produce the same transient histogram.
\smallskip
\squishlist
  \item {\bf Albedo-depth ambiguity:} A peak's height conflates radiometric fall-off and albedo. A small highly reflective (high albedo) object at a given distance will produce an equally strong peak as a larger but more diffuse object.
  \item {\bf Albedo-normal ambiguity:} Both a tilted scene patch, and a patch with low albedo will reflect less light than a head-on, high albedo one.  
  \item {\bf Orientation ambiguity:} The transient histogram is insensitive to rotational symmetry; a plane tilted at \SI{45}{\degree} clockwise or counterclockwise with respect to the $x-y$ plane will produce exactly the same transient histogram.
\squishend
\smallskip
We now present a family of techniques to recover 3D scene
information from transient histograms, beyond what can be
achieved via simple peak-finding.

\section{Plane Estimation with Transient Histograms}\label{sec:plane-est}

As pointed out by previous works~\cite{liu_planenet_2018,liu_planercnn_2019,lee_blocks-world_2021},
numerous scenes especially indoor ones can be well approximated as piecewise planar.
Here, we study the problem of recovering the
parameters of a planar scene from a captured transient.
By limiting ourselves to estimating a single plane per transient, we simplify the problem without loss of generality: one could apply our method to a small array of transients to estimate a piecewise planar scene.
Solving this problem will provide insights on estimating
more complex 3D geometry \eg, piecewise planar scenes, and
other parametric shapes.

%In our first experiment, we consider only planar scenes, as imaged by a time-resolved sensor with field of view $\theta_{fov}$, and try to estimate the plane's parameters from a captured transient. We present two different approaches that can efficiently estimate the plane parameters from the given transient.

\begin{figure*}[!t]
  \centering
  \includegraphics[width=0.95\textwidth]{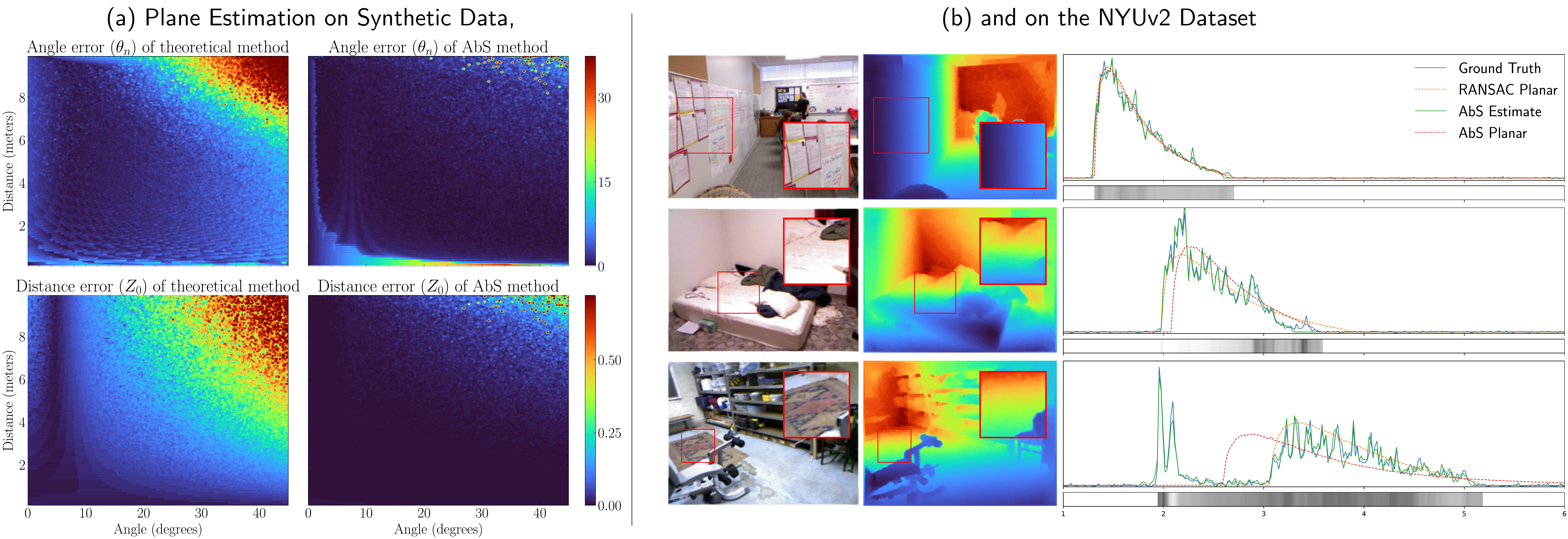}%\vspace{-1em}
  \caption{\textbf{Plane Estimation on Synthetic and NYUv2
    data.} (a) We evaluate the mean average error of
    estimated plane parameters produced by each approach
    over a wide range of parameter combinations. The first
    column corresponds to the theoretical approach, the
    second is the analysis by synthesis approach. (b) We
    show a few NYUv2 patches and their accompanying
    transients. Solid transients represent the simulated
    ground truth (blue) and the resulting transient produced
    by the AbS method (green). Dashed transients correspond
    to what we would observe if the scene was perfectly
    planar with uniform albedo with parameters as estimated
    by the RANSAC fitting process (orange), and with
    parameters estimated by the AbS approach (red). Below
    each transient, we show the average albedo per bin.
  \label{fig:plane-figs}}
  %\vspace{-1em}
\end{figure*}

\smallskip
\noindent \textbf{Plane Parametrization}: We parameterize a plane by its normal as given in spherical coordinates $\vec{n} = [1, \theta_n, \phi_n]^\intercal$, and the
distance $Z_0$, at which the plane intercepts the sensor's
optical axis (Fig.\ \ref{fig:scene-setup-radiometry}(b)). Due to
rotational symmetry, it is not possible to
recover the azimuth $\phi_n$, so we focus on
estimating $\theta_n$ and $Z_0$.

\subsection{Plane Estimation Algorithms}
\noindent \textbf{Theoretical Estimator}: For small FoVs,
$Z_0$ can be directly estimated by finding the location of
the largest peak in the transient. This estimator becomes
less accurate as the FoV increases, but in practice, this
decay can be neglected, or a better estimate can be derived
from the center of the transient's support if necessary. To
estimate $\theta_n$, we refer to Fig.\ \ref{fig:scene-setup-radiometry}(b). The
distance to a point on the plane $P$ at a viewing angle
$\gamma$, as measured from the optical axis, is given by:
\begin{equation}
||P|| = Z_0 \frac{\cos{\left(\theta_{n} \right)}}{\cos{\left(\gamma + \theta_{n} \right)}}.
    \label{eq:plane-est}
\end{equation}
With the exception of when $\theta_n$ is zero, Eq.~(\ref{eq:plane-est}) reaches its extrema at $\pm
\theta_\text{fov}/2$, corresponding to the furthest and
closest visible scene points, respectively. These extrema can be
directly estimated from the transient by detecting the
leading and lagging edges of the peak from the 1D signal.
This yields two new distances denoted $D_1, D_2$, which each
gives an estimate of $\theta_n$ by Eq.~(\ref{eq:plane-est}).
Averaging these two estimates yields our final estimate for
$\theta_n$. While such an estimator only relies on basic
peak finding primitives, it may fail for large values of
$Z_0$ and $\theta_n$ when the lagging edge of the peak 
falls outside the unambiguous depth range.

%While such an estimator relies on nothing more than basic peak finding primitives, we find that some plane parameters are not well estimated especially if the plane is severely tilted or at close/far range.

% We construct a theoretical estimator of both plane parameters that relies upon only the above derivation. This estimator relies on nothing more than basic peak finding primitives and is thus extremely fast and efficient. However, as seen in Fig.~\ref{fig:roc} some plane parameters are not well estimated especially if the plane is severely tilted or at close/far range.

\smallskip
\noindent \textbf{Analysis-by-Synthesis Algorithm}:
We introduce an analysis-by-synthesis (AbS) based estimator
that further refines the theoretical estimator.
The key idea is to directly optimize the scene parameters
($\theta_n, Z_0$) using a differentiable forward rendering
model $R(\theta_n, Z_0)$ that approximates the image
formation defined in Eq.~(\ref{eq:transient_histogram}).
This is done by discretizing the integral and replacing the
transient binning operation with a soft-binning process which uses a sharp Gaussian kernel for binning.
Specifically, given a measured transient histogram
$\widetilde \varphi = \{\widetilde\varphi_i\}_{i=1}^N$, we solve the
following optimization problem using gradient descent, with
the initial solution of $\theta_n$ and $Z_0$ given by our
theoretical approach:
\begin{equation}
\underset{\theta_n, Z_0}{\argmin} \ ||\mathcal{F}\left(R(\theta_n, Z_0)\right) - \mathcal{F}\left(\widetilde\varphi\right)||^2_2
\end{equation}
where $\mathcal{F}$ denotes the Fourier transform.  The
$\mathcal{L}_2$ norm is computed on the top $k=64$ of the
$512$ complex-valued Fourier coefficients. This is
equivalent to low-pass filtering the signal and removes high-frequency noise. Please see the supplement for details.

\subsection{Simulation Results}
\smallskip
\noindent \textbf{Quantitative Results on Synthetic Data}:
To evaluate the effectiveness of our approaches, we
simulated transients that correspond to uniform-albedo
planes with $Z_0 \in [0, 10]$ meters and $\theta_n \in [0,
45]$ degrees. For each transient, we estimated plane
parameters with the theoretical and AbS methods and compare
these to the ground truth. Results can be found in
Fig.~\ref{fig:plane-figs}(a). We observe that the AbS method
performs better than the theoretical method $87\%$ of the
time for estimating $\theta_n$ and $97\%$ for $Z_0$.

\smallskip
\noindent \textbf{Simulating Transients with RGB-D Data}: We
assume a Lambertian scene and simulate transients
produced under direct illumination using paired RGB images
and depth maps from existing RGB-D datasets. A ground truth transient
histogram is generated through Monte Carlo integration for each scene. We
sample rays emitted by the light source, march them until
they intersect a scene surface, and weight the returning signal
by scene albedo.

%\begin{figure}[!ht]
%  \centering
%  \includegraphics[width=0.9\columnwidth]{figures/r200x200bs1cs400-MAE}\vspace{-1em}
%  \caption{\textbf{Comparison of different approaches.} We evaluate the mean average error of estimated plane parameters produced by each approach. The first column corresponds to the theoretical approach, the second is the analysis by synthesis approach. We observe that the AbS method performs better than the theoretical method $87\%$ of the time for estimating $\theta_n$ and $97\%$ for $Z_0$. Also note that the AbS method can be unstable when estimating $\theta_n$ if the plane is very tilted and at close range.\label{fig:roc}}\vspace{-1em}
%\end{figure}

% AbS method is better than theoretical method at predicting theta 86.97% of the time
% AbS method is better than theoretical method at predicting z 97.29% of the time

\smallskip
\noindent \textbf{Qualitative Results on NYUv2 Dataset}:
We further test our methods on images from
NYUv2~\cite{nathan_silberman_indoor_2012} --- a well-known
RGB-D dataset. Transient histograms of local patches were
simulated, and plane fitting using RANSAC was performed on
the depth map to estimate surface normals of the patches.
Results of our methods with simulated transients as inputs
were compared against estimated surface normals, as
shown in Fig.\ \ref{fig:plane-figs}(b).

\subsection{Hardware Prototype and Results}
\begin{figure}[!t]
  \centering \includegraphics[width=0.95\columnwidth]{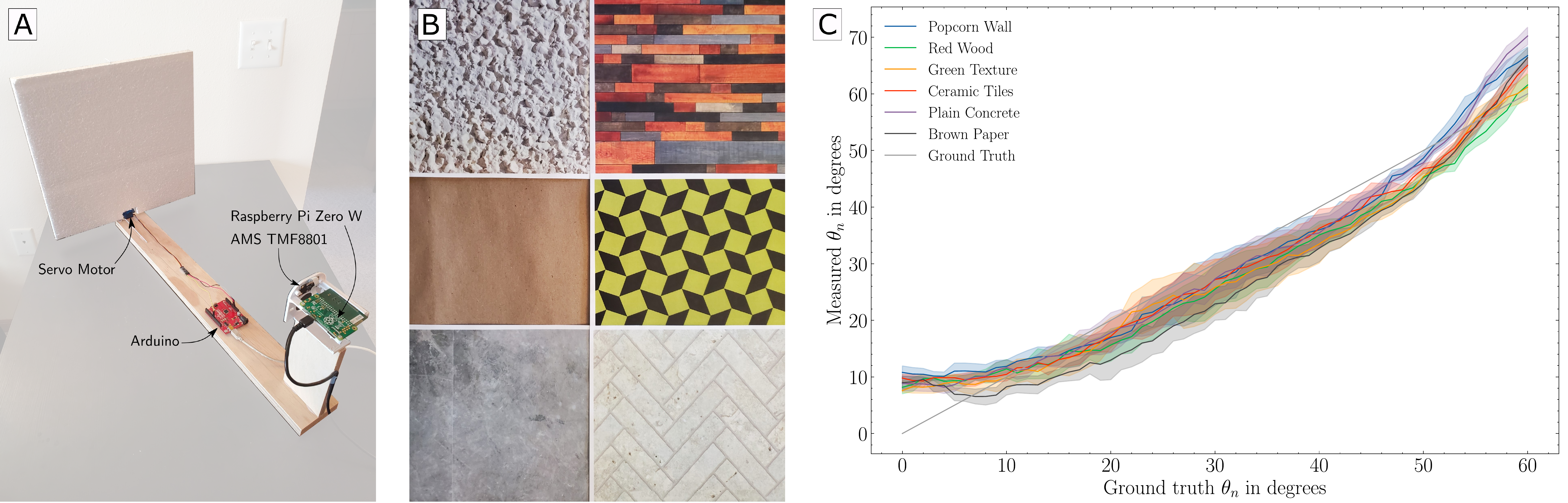}%\vspace{-0.5em}
    \caption{\textbf{Hardware prototype for plane
      estimation and results.} (a) Our low-cost prototype consists of an
      AMS TMF8801 sensor that scans a planar scene from
      different angles. (b) We captured data from a variety
      of plane textures and albedos. (c) Our method successfully estimates
      plane parameters for a wide range of plane textures.
    \label{fig:ams-setup-results}}
    %\vspace{-1.5em}
\end{figure}

We built a low-cost hardware prototype using a SPAD-based proximity sensor (AMS TMF8801, retail price $\sim$2 USD) and a Raspberry Pi Zero W mounted on a custom 3D printed mount. This sensor has a bin resolution of \SI{200}{ps} and a FoV of $\sim$20 degrees. As shown in Fig.~\ref{fig:ams-setup-results}(a) the sensor is attached to a plywood structure and scans a test plane from different angles controlled using a servo motor. The sensor and test plane are at a known distance apart. We recover the plane angle $\theta_n$ from the transient histograms. Using the servo to rotate the styrofoam backplate to all angles within $[-60^{\circ}, 60^{\circ}]$ in increments of $1^{\circ}$, we acquired 100 transients at each angle and with each of the 6 textures shown in Fig.\ \ref{fig:ams-setup-results}(b). We capture this range to acquire more data and correct miscalibrations.

%****************

\smallskip
\noindent \textbf{Results}: 
%Only our
%theoretic approach was considered in this experiment, as the
%sensor performs some unknown filtering and smoothing
%operations on the SPAD histogram that prevents the use of
%our AbS method. Further, a calibration was performed to find
%a shifting and scaling factor that cancels the filtering and
%smoothing.
Fig.~\ref{fig:ams-setup-results}(c) shows the mean $\theta_n$
estimate, as well as the standard deviation, for each of the
six textures. In line with the simulation results shown in
Fig.~\ref{fig:plane-figs}(a), our theoretical approach
gives reliable estimates with experimental data over a wide
range of angles. The estimates are inaccurate when
$\theta_n$ is small (\eg, $\le 5 ^{\circ}$).  Although our theoretical model
assumes a Lambertian plane, in practice this method is
robust to real-world albedo variations and provides 
reliable estimates even with non-Lambertian textured
surfaces.

\section{Depth Estimation via Transient Histograms}\label{sec:depth-est}

We now consider estimating depth for scenes with complex geometry and albedos. Due to the inherent rotational ambiguity of our plane estimation method, it is challenging to extend it to a small array of transient and model the scene as piecewise planar. While it is possible to estimate the remaining $\phi_n$ parameters from spatially-neighboring transients this method does not generalize well to more complex scenes, especially if they violate local planar assumptions. To accomplish this task without prior knowledge of the scene's overall shape we design a deep model to predict dense depth from a sparse set of transient histograms. We also demonstrate that the resulting depth map can be further refined with an additional RGB input image.

%Generally, we cannot parametrize the scene apriori and explicit approaches like the ones shown above for planar scenes are not feasible. In this section we look at estimating the depth of a scene in a non-parametric way, that is, without prior knowledge of the scene's overall shape. To accomplish this we devise an architecture that can predict dense depth given only a few transient histograms as inputs. We show that using a small array of transients provides better results than simply using an array of ground-truth depth values and demonstrate our results for two different grid sizes.

\smallskip
\noindent \textbf{Multiple Transients}: Recovering complex
geometry from a single transient histogram is a
severely ill-posed problem. So we propose to use
a low spatial resolution 2D array of defocused SPAD
pixels that image sub-regions of the complete FoV. Two
configurations are considered: a $4\times3$ array of SPADs,
each with a FoV of 25 degrees, and a $20\times15$ array each
with a FoV of 10 degrees. The specific fields of view are
chosen to cover the whole scene with each SPAD's FoV
overlapping slightly with its neighbors. For an output
resolution of $640\times480$, these arrays correspond to a
downsampling ratio of $160\times$ and $32\times$
respectively.

% With a high-resolution array of SPADs, each focused on a single point, we would be able to trivially recover a dense depth map albeit with a high hardware cost and high power consumption. On the other hand, we've shown we can extract more information than a single depth measurement from a transient showing that a dense array of SPADs might not be necessary.

% In practice, a single SPAD that subtends the whole scene could be used, but extracting meaningful information from it proves very difficult. Similarly, we could treat the scene as piecewise planar, estimate a $\theta_n/Z_0$ pair for each transient as we've done above, and then train a network to predict depth for the whole scene given these pairs. As we've seen in the last row of figure~\ref{fig:robustness}, for natural scenes this planar assumption does not hold everywhere and thus some estimates will be quite noisy. Instead, we opt to process the transients directly instead of relying on an intermediary representation of them. We show results for two configurations: a $4\times3$ array of SPADs, each with a FoV of 25 degrees, and a $20\times15$ array each with a FoV of 10 degrees. The specific fields of view were chosen to cover the whole scene with each SPAD's FoV overlapping slightly with its neighbors. For an output resolution of $640\times480$, these arrays correspond to a downsampling of $160\times$ and $32\times$ respectively.

\begin{figure*}[!ht]
\centering
  \includegraphics[width=1.0\textwidth]{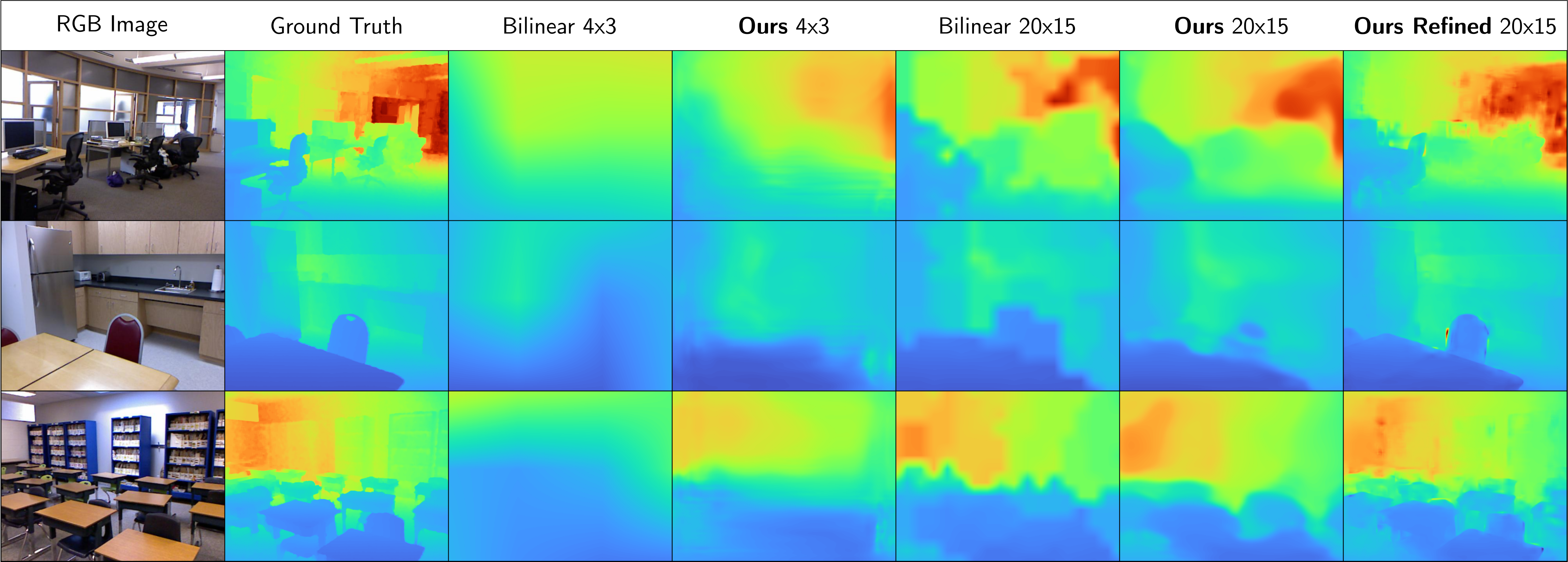}%\vspace{-0.5em}
  \caption{\textbf{Simulated Results on NYUv2 Dataset.} From
  left to right this figure shows the RGB image, ground
  truth depth, bilinear upsampling followed by our method
  using the $4\times3$ tiling, bilinear upsampling followed
  by our method using the $20\times15$ tiling, and finally
  the refined results for the $20\times15$
grid. See supplement for more results.\label{fig:qualitative-results}}
 %\vspace{-1.5em}
\end{figure*}

\subsection{Scene Depth from Transients via Deep Models}\label{sec:methods}

We now describe deep models for estimating depth from transient histograms and for refining the depth map with the guidance of an RGB image.

%In this section, we describe our proposed transient histogram-based depth estimation network and explain all of its components.

\smallskip
\noindent \textbf{Depth Estimation}: We adapt a deep convolutional neural network
architecture similar to that of recent monocular depth
estimation methods~\cite{fang_towards_2020} to perform depth estimation. The model
consists of repeating blocks of convolutional layers and
upsampling layers which are stacked until the desired output
resolution is achieved. Similar to our previous experiment
on plane estimation, we compute the Fourier transform of the
input transients and only keep the $k$ coefficients with lowest frequency. We
use $k=4$ for the $4\times 3$ grid and $k=16$ for the $20
\times 15$ grid. We train the network using the reverse
Hubert ``BerHu'' loss~\cite{verducci_robust_2007,zwald_berhu_2012}.
Details about our models and training procedures are given
in the supplement.

\smallskip
\noindent \textbf{Depth Refinement}: Estimating depth from a
sparse set of SPAD sensors is challenging due to the low
spatial resolution.  In many cases, we might have access to
an RGB image of the scene which contains high-frequency
spatial information that is lacking in the transients.
It can be used to refine our depth prediction. To accomplish
this, we combine two existing methods. The first,
FDKN~\cite{kim_deformable_2021} is trained to refine a
low-resolution depth map given an RGB image. The model was
designed to super-resolve a depth map by at most $16\times$,
beyond which there is significant performance
degradation. Directly using this network as a
post-processing step helps yet leads to noticeable artifacts
even when finetuned. To alleviate this we use the pre-trained
DPT~\cite{ranftl_vision_2021} model. Despite its low
absolute depth accuracy, DPT provides high-resolution
spatial details.

Specifically, the depth map produced by the FDKN network is
used as a guidance image which determines how the depth map
from DPT should be deformed. On a tile-by-tile basis, we
compute the scale and shift that minimizes the
$\mathcal{L}_2$ loss between the two depth maps. Once this
transformation is interpolated over the whole depth map and
applied to the DPT prediction, we get our final result: a
depth map with greater spatial detail and higher
accuracy than MDE.

\begin{table*}[!t]
    \centering
        \caption{\textbf{Results on NYUv2 Benchmark.} We show numerical results for our approaches, baselines, and well-known MDE methods. While MDE approaches can produce detailed depth maps, their absolute depth accuracy (in the lower $\delta$ metrics) is at par with our $4\times3$ grid. Our approach produces more accurate depth maps overall. 
        % they often have large errors in absolute depth which have about the same $\delta < 1.05^1$ accuracy as what is achievable with a small $4\times3$ grid. Our approach produces more accurate depth maps overall.
        % Some works~\cite{ranftl_towards_2020, ranftl_vision_2021} avoid this issue by simply predicting depth up to a scale and shift, yet for any real application, this missing scale and shift are key.
    }
    \resizebox{\columnwidth}{!}{
    \begin{tabular}{r|r|cccccc|cHHHccH}\hline
        \multicolumn{1}{c|}{Grid Size} & \multicolumn{1}{c|}{Method} & $\delta < 1.05^1\uparrow$ & $\delta < 1.05^2\uparrow$ & $\delta < 1.05^3\uparrow$ & $\delta < 1.25^1\uparrow$ & $\delta < 1.25^2\uparrow$ & $\delta < 1.25^3\uparrow$ & $Log_{10}\downarrow$ & $mae\downarrow$ & $med acc\downarrow$ & $mse\downarrow$ & $AbsRel\downarrow$ & $RMSE\downarrow$ & $sq rel\downarrow$ \\ \hline \hline

        % berhu_4_fftcartesian_4x3f25, berhu_base_z_4x3f25, UpsampleEstimator
        \multirow{3}{*}{$4 \times 3$} & Ours                      & \underline{0.335} & \textbf{0.577}    & \textbf{0.724}    & \textbf{0.845}    & \textbf{0.953}    & \textbf{0.981}    & \textbf{0.068}    & \textbf{0.354}    & \textbf{1.085}    & \textbf{0.420}    & \textbf{0.126}    & \textbf{0.604}    & \textbf{0.112}    \\
        & Baseline                                                & \textbf{0.340}    & \underline{0.569} & \underline{0.709} & \underline{0.824} & \underline{0.934} & \underline{0.967} & 0.171             & \underline{0.385} & \underline{1.087} & \underline{0.483} & \underline{0.147} & \underline{0.652} & \underline{0.153} \\
        & Bilinear                                                & 0.285             & 0.466             & 0.588             & 0.715             & 0.886             & 0.951             & \underline{0.083} & 0.527             & 1.127             & 0.886             & 0.169             & 0.856             & 0.197             \\
        \hline

        % berhu16fftcartesian20x15f10, basez20x15f10, UpsampleEstimator
        \multirow{4}{*}{$20 \times 15$} & Ours          & \textbf{0.624}    & \textbf{0.809}    & \textbf{0.880}    & \textbf{0.929}    & \textbf{0.976}    & \textbf{0.989}    & \underline{0.060} & \textbf{0.199}    & \textbf{1.035}    & \textbf{0.189}    & \textbf{0.073}    & \textbf{0.409}    & \textbf{0.059}    \\
        & Baseline                                      & 0.576             & \underline{0.786} & \underline{0.867} & \underline{0.923} & \underline{0.973} & \underline{0.988} & 0.066             & \underline{0.218} & 1.041             & \underline{0.239} & 0.084             & \underline{0.450} & 0.113             \\
        & Bilinear                                      & \underline{0.583} & 0.763             & 0.840             & 0.899             & 0.963             & 0.985             & \textbf{0.038}    & 0.246             & \underline{1.039} & 0.299             & \underline{0.081} & 0.498             & \underline{0.069} \\

        \cline{2-15}
        & Ours Refined                                  & \textbf{0.707}    & \textbf{0.865}    & \textbf{0.924}    & \textbf{0.961}    & \textbf{0.990}    & \textbf{0.996}    & \textbf{0.024}    & \textbf{0.139}    & \textbf{1.026}    & \textbf{0.091}    & \textbf{0.053}    & \textbf{0.287}      & \textbf{0.026}    \\
        \hline \hline

        \multirow{4}{*}{MDE} & DORN~\cite{fu_deep_2018} & \textbf{0.394}    & \underline{0.602} & 0.731             & 0.846             & 0.954             & 0.983             & 0.053             & -                 & -                 & -                 & 0.120             & 0.501             & -                 \\
        & DenseDepth~\cite{alhashim_high_2019}          & 0.311             & 0.548             & 0.706             & 0.847             & 0.973             & 0.994             & 0.053             & -                 & -                 & -                 & 0.123             & 0.461             & -                 \\
        & BTS-DenseNet~\cite{lee_big_2020}              & \underline{0.357} & \textbf{0.607}    & \underline{0.764} & \underline{0.885} & \underline{0.978} & \underline{0.994} & \underline{0.047} & -                 & -                 & -                 & \underline{0.110} & \underline{0.392} & -                 \\
        & DPT~\cite{ranftl_vision_2021}                 & 0.326             & 0.595             & \textbf{0.767}    & \textbf{0.904}    & \textbf{0.988}    & \textbf{0.998}    & \textbf{0.045}    & -                 & -                 & -                 & \textbf{0.109}    & \textbf{0.357}    & -                 \\
        \hline

        \end{tabular}
    }
    % \vspace{0.5em}

    %\vspace{-1em}
    \label{tab:result_table}
\end{table*}

\subsection{Simulation Results}\label{sec:results}

%In this section, we show simulation results on the NYUv2 indoor dataset~\cite{nathan_silberman_indoor_2012}. We first show qualitative results and how these compare upsampling. Then we quantitatively evaluate our approach and compare it with other methods including some recent MDE techniques.

\smallskip
\noindent\textbf{Dataset:} We use the standard test/train split of the widely used NYUv2 dataset \cite{nathan_silberman_indoor_2012} which is a large indoor dataset with dense ground truths.

\smallskip
\noindent\textbf{Evaluation Metrics:} To quantitatively evaluate our results, we compare our approach to existing methods using the standard metrics used in prior work~\cite{eigen_depth_2014}, including Absolute Relative Error (AbsRel), Root Mean Squared Error (RMSE), Average Log Error (Log10), and Threshold Accuracy ($\delta < thr$). See supplement.

Existing literature focuses on $\{1.25, 1.25^2, 1.25^3\}$ thresholds that correspond to $25\%$, $56\%$, and $95\%$ depth error. However, we believe that many real world applications such as robot navigation and obstacle avoidance need stronger accuracy guarantees. To better quantify the gap between LiDARs and MDE-based methods, we consider three stricter thresholds $\{1.05,1.05^2, 1.05^3\}$ that correspond to $5\%$, $10\%$ and $16\%$ error.

\smallskip
\noindent\textbf{Baselines:} A simple baseline uses bilinear upsampling of the tiled depth as computed via peak-finding. We also consider a stronger baseline that uses a deep network to super-resolve depth maps at each tile. We compare with recent MDE methods~\cite{fu_deep_2018,alhashim_high_2019,lee_big_2020,ranftl_vision_2021} for which some metrics were re-computed using the pre-trained models as these were not published in the original papers.

\smallskip
\noindent\textbf{Qualitative Results:} Fig.~\ref{fig:qualitative-results} compares our method against various baselines. 
Observe that in the first and last rows, our network can extract more information from a transient: farther scene depths that are missing in a bilinear-upsampled depth maps are visible with our method.
This effect is particularly noticeable for the smaller $4\times 3$ grid. The last column shows results of our
depth-refinement method which adds more spatial details.

\smallskip
\noindent\textbf{Quantitative Results:}
Table~\ref{tab:result_table} presents our main results. As seen in the lower $\delta$ metrics, our method provides the most benefits using small grids. We observe a $5\%$ accuracy increase over bilinear in the lowest $\delta$ metric for the small grid whereas it only boosts it by $4.1\%$ in the larger grid. Moreover, observe that our refinement step increases the $5\%$-accuracy metric by more than $8\%$, and reaches nearly double the accuracy of some MDE approaches.

\smallskip
\noindent \textbf{Energy, latency, and cost:} Table~\ref{tab:ressources} shows a comparison of our method ($4 \times 3$) using a (simulated) low-cost SPAD vs. a conventional RGB camera in terms of power consumption, bandwidth, compute cost, and depth accuracy. Our method generates similar depth quality as MDE approaches while consuming $\nicefrac{1}{10}^\text{th}$ the power, 2400$\times$ less bandwidth, and orders of magnitude less compute. We expect such low-resolution ($4\times 3$) SPAD arrays will be significantly cheaper than high resolution RGB and ToF camera modules. Table~\ref{tab:ressources} does not include the optional RGB refinement step which will consume additional resources.

%By treating transient histograms as a \emph{primitive scene representation} (which can be optionally augmented with RGB images), we benefit in terms of bandwidth, power, and compute costs.

\begin{table}[!t]
    \centering
    \caption{\textbf{Sensing power and compute costs.} Power estimates are based on an AMS TMF8828 proximity sensor and a typical smartphone camera. Timing analysis is conducted on an Intel Core i7-9700K CPU and Nvidia RTX 2080 SUPER GPU with an input resolution of $640 \times 480$ and averaged over 300 runs.}
    %This table shows the sensing peak-power, per-frame bandwidth and compute requirements (measured using \href{https://github.com/facebookresearch/fvcore/blob/main/docs/flop_count.md}{fvcore}), number of parameters and inference time for various methods. 
   % and averaged over 300 runs. We used images and depth maps of size $640 \times 480$ for above MDE comparisons.
    \begin{tabular}{r|cccccc}\hline
        Method & Peak Power & Bandwidth & GFLOPS & \#Params & Time & $\delta < 1.25^3\uparrow$ \\ \hline \hline

        Ours $4 \times 3$ & $141$ mW & $384$ B & $0.811$ & $28.6$ K & $2.77$ ms & $0.981$ \\
        % Ours $20 \times 15$ & $141$ mW & $37.5$ kB & $0.807$ & $21.8$ K & $2.48$ ms \\
        DORN [9] & $1.98$ W & $900$ kB & $389.5$ & $162.7$ M & $112.0$ ms & $0.983$\\
        DPT [30] & $1.98$ W & $900$ kB & $280.3$ & $120.7$ M & $81.4$ ms & $ 0.998$ \\
        \hline
        % Smartphone power estimate is based on typical current draw as reported here: https://source.android.com/devices/tech/power/values
        % And a bus voltage of 3.3v
    \end{tabular}
    
    \vspace{0.5em}
    
    % \vspace{-1.5em}
    \label{tab:ressources}
\end{table}

\subsection{Hardware Prototype and Experiment Results}\label{subsec:lab-setup}

\begin{figure}[!t]
  \centering \includegraphics[width=0.95\columnwidth]{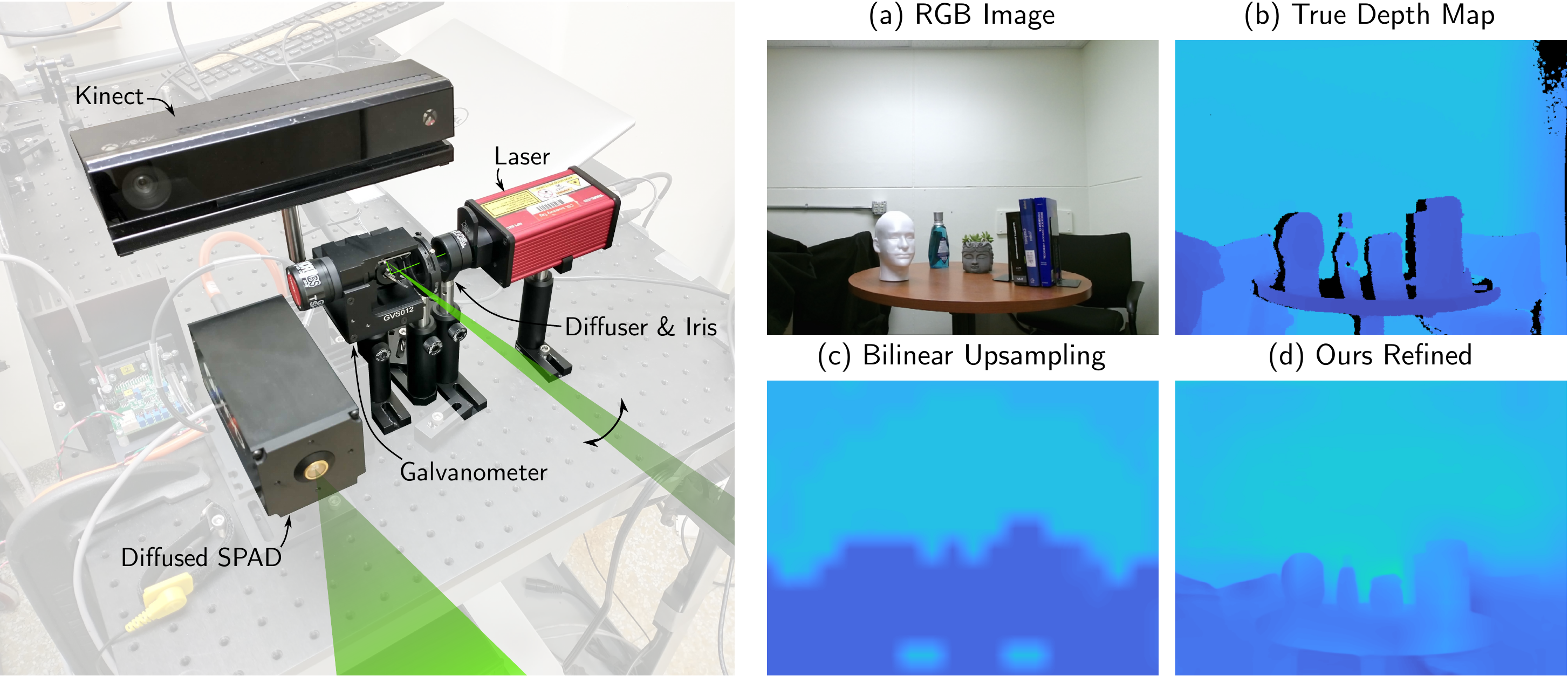}%\vspace{-1em}
    \caption{\textbf{Hardware prototype and experiment results.} Our hardware
    prototype (left) scans a $20\times15$ grid in the scene with a
    diffused laser spot. A lens-less SPAD pixel
    captures transient histograms.
    (a) We imaged a table-top scene with a wide range of
    albedos and textured objects. (b) The true depth map
    captured using a Kinect v2. (c) Simple peak-finding-based
    depth map provides no depth details. (d) Using the
    RGB image and an MDE model,
    our method generates a higher resolution depth map.
    \label{fig:lab-setup-results}}
    % \vspace{-0.2in}
\end{figure}

Our lab cart hardware prototype (Fig.~\ref{fig:lab-setup-results}) was designed to be similar in operation to a low-cost off-the-shelf AMS TMF8828 proximity sensor. However, this setup provides greater flexibility with reprogramming and alignment with the RGB-D camera, and allows evaluating different scanning patterns. Our setup consists of a pulsed laser (Thorlabs NPL52C) with a \SI{6}{\nano\second} pulse width, \SI{1.2}{W} peak power, and \SI{40}{\kilo\hertz} repetition frequency. The laser spot is diffused and shaped into a $\SI{10}{\degree}$ cone using a diffuser (Thorlabs ED1-C20) and an adjustable circular iris. The detector is a lens-less single-pixel SPAD (MPD InGaAs Fast Gated SPAD) operated in gated acquisition mode with a dead-time of \SI{1}{\us}. The FoV of the SPAD pixel covers the whole scene. A 2-axis galvanometer (Thorlabs GVS012) scans a $20\times 15$ grid that covers the FoV. In practice, this can be replaced with a low-resolution multi-pixel SPAD array. Photon timestamps are acquired using a time-correlated single-photon counting (TCSPC) system and histograms are constructed offline. A Microsoft Kinect v2 RGB-D camera provides ground truth intensity and depth maps.

\smallskip
\noindent \textbf{Results:}  Fig.~\ref{fig:lab-setup-results} (right panel) also shows results on an indoor
table-top scene using our setup. This scene is challenging due to several objects with varying reflectances and sharp depth edges. Using bilinear upsampling from the transient peaks results in jagged edges and overall loss of detail. The proposed methods can recover fine details and accurate depths with as few as
$20\times15$ transients. For more results and comparisons, please refer to the supplementary material. 

\section{Limitations and Discussion}\label{sec:limitations}

\smallskip
\noindent \textbf{Bottlenecks and Availability:}
% Bandwidth and Frame rates + Sensor Availability
Proximity sensors containing small arrays of SPAD
pixels are already commonplace in consumer electronics.
However, they only output a single depth
measurement. Some off-the-shelf sensors provide limited access to
low-resolution pre-processed transients.
Bandwidth bottlenecks between the sensor and
processing unit limit the rate at which transients can be read out.
For the methods described in this paper to become widespread,
sensor manufacturers need
to address communication bottlenecks, and document and advertise
low-level APIs that allow direct access to transients.

\smallskip
\noindent \textbf{Future Outlook:}
Faster communication protocols and on-chip compression methods will
enable capturing transient histograms at high frame rates
using commodity hardware.
These can not only be used to determine scene
geometry but also help resolve objects in low-light, detect
fast-moving targets, and detect subtle
scene motion (e.g. heartbeat or acoustic
vibrations). Transient histograms can be treated as a
primitive scene representation complementary to RGB images
for time-critical and resource-constrained applications. \smallskip

\noindent {\bf Acknowledgments:} This research was supported in part by the NSF CAREER award 1943149, NSF award CNS-2107060 and Intel-NSF award CNS-2003129. We thank Talha Sultan for help with data acquisition.

%%%%%%%%% REFERENCES
\clearpage
%\nocite{*} % Remove this!
\bibliographystyle{splncs04}
\bibliography{HistoVision}

% \clearpage
%\onecolumn
% \input{supplement}

\end{document}